\documentclass[
]{ceurart}

\usepackage{listings}
\usepackage{float}
\usepackage{placeins}
\usepackage{booktabs}

\usepackage{caption} 
\usepackage{tabularx}                     
\usepackage{listings}    
\usepackage{hyperref}    

\lstdefinestyle{promptjson}{
  language=Java,
  morekeywords={true,false,null},
  basicstyle=\ttfamily\footnotesize,
  breaklines=true,
  breakatwhitespace=false,
  columns=flexible,
  frame=single,
  showstringspaces=false,
  numbers=none
}

\newcolumntype{Y}{>{\ttfamily\raggedright\arraybackslash}X}

\begin{document}

\copyrightyear{2025}
\copyrightclause{Copyright for this paper by its authors.
  Use permitted under Creative Commons License Attribution 4.0
  International (CC BY 4.0).}

\conference{CLEF 2025 Working Notes, 9 -- 12 September 2025, Madrid, Spain}

\title{DS@GT at CheckThat! 2025: A Simple Retrieval-First, LLM-Backed Framework for Claim Normalization}

\title[mode=sub]{Notebook for the <CheckThat!> Lab at CLEF 2025}

\author[1]{Aleksandar Pramov}[
orcid=0009-0005-9049-1337,
email=apramov3@gatech.edu,
]\cortext[1]{Corresponding author.}\cormark[1]
\author[1]{Jiangqin Ma}[
orcid=0009-0006-0058-6349,
email=jma416@gatech.edu,
]
\author[1]{Bina Patel}[
orcid=0009-0002-7815-6331,
email=bpatel80@gatech.edu,
]
 
\address[1]{Georgia Institute of Technology, North Ave NW, Atlanta, GA 30332}
 
\begin{abstract}
Claim normalization is an integral part of any automatic fact-check verification system. It parses the typically noisy claim data, such as social media posts into normalized claims, which are then fed into downstream veracity classification tasks. The CheckThat! 2025 Task 2 focuses specifically on claim normalization and spans 20 languages under monolingual and zero-shot conditions.  Our proposed solution consists of a lightweight \emph{retrieval-first, LLM-backed} pipeline, in which we either dynamically prompt a GPT-4o-mini with in-context examples, or retrieve the closest normalization from the train dataset directly. On the official test set, the system ranks near the top for most monolingual tracks, achieving first place in 7 out of of the 13 languages. In contrast, the system underperforms in the zero-shot setting, highlighting the limitation of the proposed solution.
\end{abstract}

\begin{keywords}
  Claim normalization \sep
  Large language models \sep
  BERTScore \sep
  Sentence Transformers \sep
  Multilingual NLP
\end{keywords}

\maketitle

\section{Introduction}

The CheckThat! lab studies critical steps of the pipeline of automatic claim verification systems \cite{CheckThat:ECIR2025, clef-checkthat:2025-lncs}. Such systems typically consist of several high-level steps: (i) establishing check-worthy claims, (ii) parsing and normalization of said claims, (iii) retrieving relevant evidence and (iv) veracity classification.

Subtask 2 of the CheckThat! 2025 edition  focuses on the second stage of this process \cite{clef-checkthat:2025:task2}. In particular, the task is to normalize social media posts into simpler and cleaner claims that make it easier for the downstream tasks to perform.  The task is to perform claim normalization for 20 languages of different language families, some of which are labeled with training data (\emph{monolingual} setting) while others are in a \emph{zero-shot} setting without any training data. The normalized claims are compared to gold normalizations (created by human fact checkers) by using the METEOR score \cite{banerjee2005meteor}.

In this paper, we first perform a thorough EDA which reveals several challenges in the dataset, among which low-context cases of post-normalization pairs, as well as high semantic overlap between posts in the \emph{train}, \emph{dev} and \emph{test} datasets. 
Following our insights from the EDA, we present a lightweight yet competitive approach to the CheckThat! 2025 Task 2. For every test post, we first retrieve its closest match from the pooled train–dev corpus using language-appropriate sentence transformers; if the cosine similarity exceeds a tuned threshold $k$, we simply reuse the best matching normalization from the available train or dev dataset. Otherwise, we fall back to a few-shot prompt for GPT-4o-mini \cite{OpenAI_GPT4_2023} that is dynamically populated with the top-3 most similar train–dev examples. 

This retrieval-first, LLM-backed design (i) avoids training a potentially over-fitting Seq2Seq model, (ii) handles the low-context cases revealed by our EDA, and (iii) transfers, albeit with a lower performance, in the zero-shot languages by switching to a fixed English prompt. This minimal pipeline performs well for many monolingual tracks across diverse languages, but performs subpar in a zero-shot context.

\section{Related Work}
Prevalent use and exposure to social media has exposed users to misleading claims originating from these posts.  Moreover, these posts tend to be laden with noise and other extraneous information that is not relevant to the main claim presented in the post. Previous work to analyze social media claims and extrapolate the main information from them explores the idea of claim check worthiness estimation, claim span identification, as fitting in the larger context of automatic claim verification \cite{sundriyal2022empowering, sundriyal-etal-2023-chaos, hassan2017claimbuster}.  
Claim span identification (CSI) can be performed as a precursory step in a fact- checking pipeline to automatically identify and extract detailed text spans from a larger text corpus to verify. Subsequently, claim normalization involves the process of text summarization once a candidate claim has been identified.
 
Claim Normalization, ClaimNorm, presents a framework for extracting relevant claims from convoluted and noisy social media posts into normalized claims. A normalized claim is information derived from social media posts in a succinct, understandable form that highlights the central idea that is made in the post and was first introduced and discussed in \cite{sundriyal-etal-2023-chaos} and serves as a basis for our work here.

In particular \cite{sundriyal-etal-2023-chaos} shows that careful in-context examples, Chain-of-Thought (CoT) reasoning, coupled with reverse-check worthiness instructions for LLMs delivers performance better than traditional sequence-to-sequence models. This process provides a degree of certainty to determine the central claims of a post, allowing for effective claim normalization. More importantly, it presents a framework that mimics how professional fact checkers fact check their work. 

Our framework draws inspiration from the Check-worthiness Aware Claim Normalization (CACN) method outlined in \cite{sundriyal-etal-2023-chaos} and combines it with a simple retrieval step driven by the insights revealed in Exploratory Data Analysis (EDA).

\section{Exploratory Data Analysis}
The task is defined as a generation problem across 20 languages, with the test set covering a wide range of linguistic and cultural contexts, including English, Arabic, Bengali, Czech, German, Greek, French, Hindi, Korean, Marathi, Indonesian, Dutch, Punjabi, Polish, Portuguese, Romanian, Spanish, Tamil, Telugu, and Thai. Due to the availability of training data and pre-processing tools, we conduct our EDA primarily on the English training dataset. This allows us to investigate structural patterns, noise characteristics, and semantic compression dynamics in a representative subset of the data.

\subsection{Corpus composition \& statistics} 
The English training dataset consists of 11,374 examples, each containing two fields: a raw social media \texttt{post} and its corresponding \texttt{normalized claim}. These posts are typically noisy and unstructured, often containing emojis, hashtags, URLs, or repetitive phrasing. The normalized claims are concise factual statements derived from the posts, suitable for fact-checking or downstream knowledge extraction. This dataset provides a strong foundation for understanding the nature of informal user-generated content and the extent of transformation required to produce verifiable claims.

To understand the textual complexity of the dataset, we analyze the word count distributions of both posts and normalized claims. The posts are significantly longer and more variable in length, often exceeding 90 tokens due to informal language, repetitive structures, and off-topic content. Moreover, a multitude of posts have the same repeated text exactly three times within the same post (a plausible hypothesis for this is that this is due to a data error at the retrieval of the posts). In contrast, normalized claims are much shorter, typically under 20 tokens. This observation underscores the need for models that can perform both semantic understanding and linguistic compression.
  
We further examine the structural characteristics of posts by identifying the presence of emojis, hashtags, and URLs. Hashtags are the most common, indicating widespread topical tagging and social signaling. Emojis are also frequently used, reflecting the expressive and informal tone of social media content. URLs appear less often, but frequently link to external evidence. These findings are summarized in Table~\ref{tab:structure_counts}, reinforcing the multimodal nature of the data and the importance of robust pre-processing strategies.

\begin{table}[h]
\centering
\caption{Number of posts containing emojis, hashtags, and URLs.}
\label{tab:structure_counts}
\begin{tabular}{lc}
\toprule
\textbf{Feature} & \textbf{Number of Posts} \\
\midrule
Hashtags & 2,013 \\
Emojis   & 1,682 \\
URLs     &   627 \\
\bottomrule
\end{tabular}
\end{table}

\begin{figure}
\centerline{\includegraphics[width=0.9\textwidth]{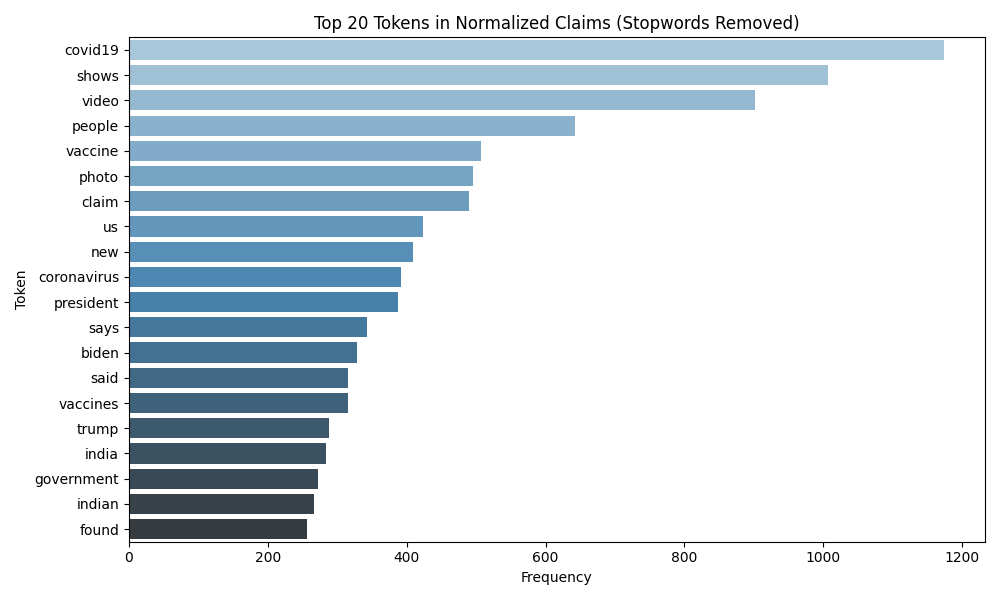}}
\caption{Top 20 most frequent word-level tokens in normalized claims (stopwords removed).}
\label{fig3}
\end{figure}

To identify common topics and patterns in normalized claims, we analyze token frequencies after removing English stopwords using a word-level tokenizer (whitespace-based, with punctuation removed). The most frequent terms included \textit{“covid19”}, \textit{“video”}, \textit{“shows”}, and \textit{“president”}, suggesting a focus on politically and visually grounded misinformation. This analysis informs downstream tasks such as topic modeling and schema-guided generation. Results are shown in Fig.~\ref{fig3}.

We also compute average token counts using the same word-level tokenization approach, defined as lowercase words split on whitespace with punctuation removed. Table~\ref{tab:token_stats} summarizes the results for the English training and development splits. On average, posts contain over 90 tokens, while normalized claims are much shorter (around 14 tokens), underscoring the extent of linguistic compression involved in the task.

\begin{table}[h]
\centering
\caption{Average word-level token count in original posts and normalized claims.}
\label{tab:token_stats}
\begin{tabular}{lcc}
\toprule
\textbf{Split} & \textbf{Avg. Tokens (Post)} & \textbf{Avg. Tokens (Claim)} \\
\midrule
Train   & 93.87 & 14.43 \\
Dev     & 94.25 & 14.14 \\
Overall & 94.06 & 14.28 \\
\bottomrule
\end{tabular}
\end{table}

\subsection{Data quality \& overlap analysis}
Apart from its textual complexity, the English dataset also exhibits other structural data quality challenges (\textbf{C1}-\textbf{C3}) which are noteworthy and build the reasoning for our subsequent modeling pipeline. These key challenges are:

\begin{description}
    \item[(C1) Mixed languages:] Some posts are in a language different than English (or mixed)
    \item[(C2) Missing context:] Many posts (i.e. non-normalized claims) have words that have no semantic match in the normalized claim.  
    \item[(C3) Semantic post overlap between the pooled dev \& train and test datasets:] Some non-normalized posts in the train dataset have either the same or a very close counterpart in the dev and/or test dataset.
\end{description}
While (\textbf{C1}) can be alleviated by either machine translating the entire claim or by ignoring it altogether, (\textbf{C2}) and (\textbf{C3}) have a greater impact on the choice of the modeling step. We give examples and discuss the implications below.

\subsubsection*{Missing context}
A particular example of (\textbf{C2}) is one post shown below - Nr. 17 from the English train dataset. Over half of the underlying post is in Hindi but here we reproduce entirely in English via machine-translation, to illustrate the point:
\begin{description} 
\item[\normalfont \emph{Original Post}:]
``Share the video as much as possible Jai Shri Ram Share the video as much as possible Jai Shri Ram Share the video as much as possible Jai Shri Ram who feeds no friend no man to love him All guilt is he who eats
alone (so eat together give and share)"
\item[\normalfont \emph{Normalized Post}:]
 ``Radio in Spain is broadcasting sacred healing verses in Sanskrit during the COVID-19 pandemic."
\end{description}
 
Apart from the mixed languages, which can be easily fixed by an automatic translation tool, the original claim (even post translation) does not contain any words, or synonyms, of the terms in the normalized claim (e.g. radio, Spain, broadcasting, Covid-19 etc.) The normalized claim misses the context the human annotator used to produce the normalization of the post. This would be an example of a 'low-context' normalization instance. It would be very difficult for a model which has just the original claim as an input to produce anything close to the normalized claim in this instance.

\begin{figure}[h]  
  \centering
  \includegraphics[width=0.6\textwidth, height=0.5\textwidth]{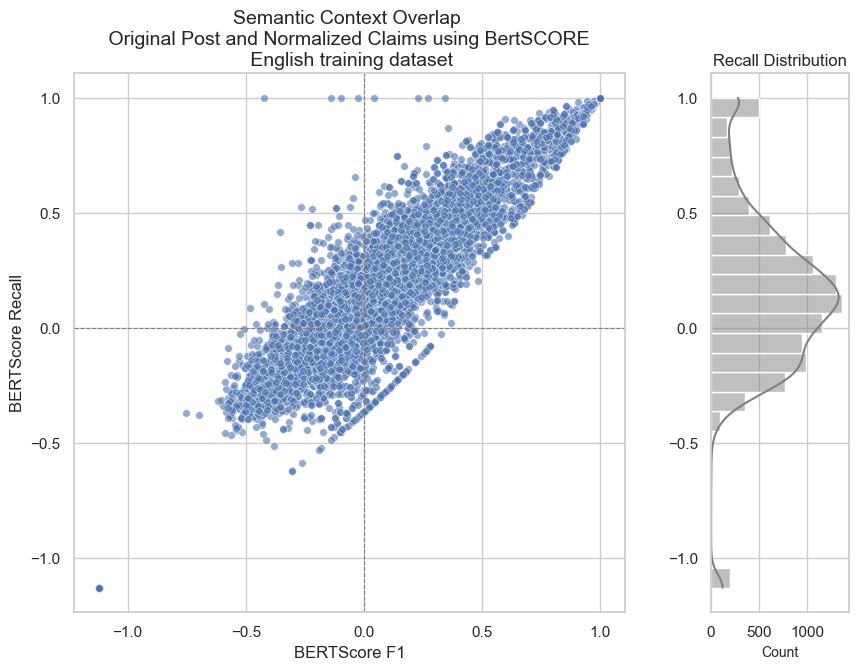} %
  \caption{BERTScore Recall vs. F1 for English train set. Low recall values (left side of the plot) indicate many normalized claims are not semantically grounded in the original post.}
  \label{fig:bertscore}
\end{figure}

To illustrate this is not an isolated case, we use Bertscore, an automatic evaluation metric for text generation \cite{zhang2019bertscore}. Unlike other metrics Bertscore computes token similarity using contextual embeddings. It's two components (Precision and Recall) match the semantic overlap between the candidate and the reference set and build the final score as a mixture of the two. Unlike Precision, which measures how much of the original post is reused, Recall captures how much of the normalized claim can be traced back to the original post (semantically). A low Recall score suggests key concepts in the normalized claim (e.g., "radio," "Spain," or "COVID-19") are not semantically present in the original post. This serves as an indicator of missing context; information the annotator relied on but which the model would not have access to. The normalized posts with high recall will have at the very least synonyms in the non-normalized posts, which in turn would make the inference task possible, unlike the aforementioned example of Post Nr. 17 above.

Figure \ref{fig:bertscore} visualizes this for the English train set. We observe that Recall scores are generally low (mean of 0.15), indicating many normalized claims are not semantically recoverable from their respective posts. This empirically supports (\textbf{C2}), confirming the presence of a substantial number of “low-context” normalization instances.

\subsubsection*{Semantic overlap in the posts between the pooled dev \& train and test datasets}
Another challenge in the data (\textbf{C3}), with effects on the modeling, comes from the fact for many posts, there is a  significant overlap between the posts in the \emph{train}, \emph{dev} (used by the organizers as a validation set in the example notebook) and \emph{test} sets.  Naturally, for such posts, the \emph{train} and \emph{dev} datasets already contain the normalization and thus at inference time, the normalization would have leaked. Here is an example of the second post in the test set:

\begin{description}
    \item[\normalfont \emph{Non-normalized claim in the dev \& train pooled set}:] ``Pence unfollowed Trump, and then changed his banner picture to Biden and Kamala. He’s outta there Pence unfollowed Trump, and then changed his banner picture to Biden and Kamala. He’s outta there Pence unfollowed Trump, and then changed his banner picture to Biden and Kamala. He’s outta there Mike Pence @Mike Pence Vice President of the United States Donald Trump.comJoined February 2009 48 Following 6.1M Followers Followed by Lauren Chen, Josh, and 8 others you follow 000 Follow"
    \item[\normalfont \emph{Non-normalized claim in the test set}:]
    ``Pence unfollowed Trump, and then changed his banner picture to Biden and Kamala. He’s outta there Pence unfollowed Trump, and then changed his banner picture to Biden and Kamala. He’s outta there Pence unfollowed Trump, and then changed his banner picture to Biden and Kamala. He’s outta there Mike Pence @Mike Pence Vice President of the United States Donald Trump.comJoined February 2009 48 Following 6.1M Followers Followed by Lauren Chen, Josh, and 8 others you follow 000 Follow"
    \item[\normalfont \emph{Normalized claim in the dev \& train pooled set}:]``Says Mike Pence changed “his Twitter banner photo to Biden and Harris."
    \item[\normalfont \emph{Normalized claim in the test set (gold output)}:] ``Vice President Mike Pence unfollowed the president and changed his Twitter banner to an image of Joe Biden and Kamala Harris"
\end{description}
We do not have access to the gold normalization of the test set at inference time. However, if the underlying non-normalized claim has perfect match between the datasets, then there would be significant overlap between the normalizations too. And since we do know one of the normalizations (from the test pooled dev \& train set), then this bears information about the gold normalization as well.

While we did not receive further clarification on this by the organizers, it is left for speculation whether this constitutes a data leak or reflects the reality of the posts, where multiple very similar (or completely identical) posts can appear from different posters by re-sharing them. This has real implications for the choice of model and the training. Seq2seq models on the training data and using the \emph{dev} dataset for validation will give distorted results. Indeed, we observed that when we let the epochs increase in the baseline \emph{seq2seq} model provided in the starter notebook, our validation metric kept improving, signaling overfitting - which is unsurprising given the overlap between training and dev datasets.

To investigate this further for the English dataset, we embed (using \emph{sentence-transformers/msmarco-distilbert-base-v3} \cite{lhoest-etal-2022-hub, wolf-etal-2020-transformers}) all of the \emph{test set}, as well the pooled \emph{dev set} \& \emph{train set} of posts (non-normalized claims) and compute the cosine similarity score between each post in the test set and all other posts in the pooled test \& dev datasets. The following figure shows the histogram of the highest similarity that was found per each test post:

\begin{figure}[h]  
  \centering
  \includegraphics[width=0.6\textwidth, height=0.5\textwidth]{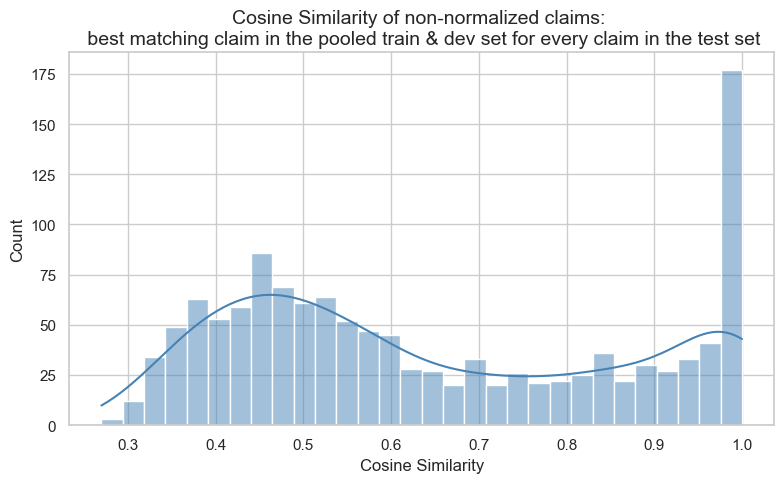} %
  \caption{Cosine similarity: Histogram of the cosine similarity between every claim in the test set vs. the best-matching (i.e. with the highest similarity) claim from the pooled train \& dev sets. In multiple instances, the similarity is very high.}
  \label{fig:simscore-distro}
\end{figure}

It is evident for multiple instances the cosine similarity is very high. In such a case (like the post above which has a complete overlap between test and pooled dev \& train sets) the best possible inference that a model can take for the normalization in the test set is the already accessible normalization from said instance in the dev \& train set.

In summary, observations \textbf{C1}, \textbf{C2}, \textbf{C3} revealed by our EDA, motivate a retrieval-centric design:

\begin{itemize}
    \item Because many test posts have near-duplicates in the pooled train–dev corpus (C3), a nearest-neighbour lookup will already return an adequate normalization for a sizable slice of the data, eliminating the need for generation.
    \item  For the remaining posts, retrieval still pays off: the top-k neighbours provide the contextual clues that are absent in low-recall, “missing-context’’ cases (\textbf{C2}) and naturally bridge code-mixed or translated fragments (\textbf{C1}).
\end{itemize}

\section{Methodology}
  
Motivated by the structural insights from (\textbf{C2}) and (\textbf{C3}), we design a simple retrieval-based normalization pipeline that searches for the most similar example from the pooled train and dev sets for each test post. This approach sidesteps the need to train a seq2seq model, which would (a) likely overfit the validation set due to substantial overlap with the training data, and (b) struggle to generate correct normalizations in low-recall scenarios, where critical context is missing from the original post.

\begin{figure}[h]  
  \centering
  \includegraphics[width=0.8\textwidth, height=1.1\textwidth]{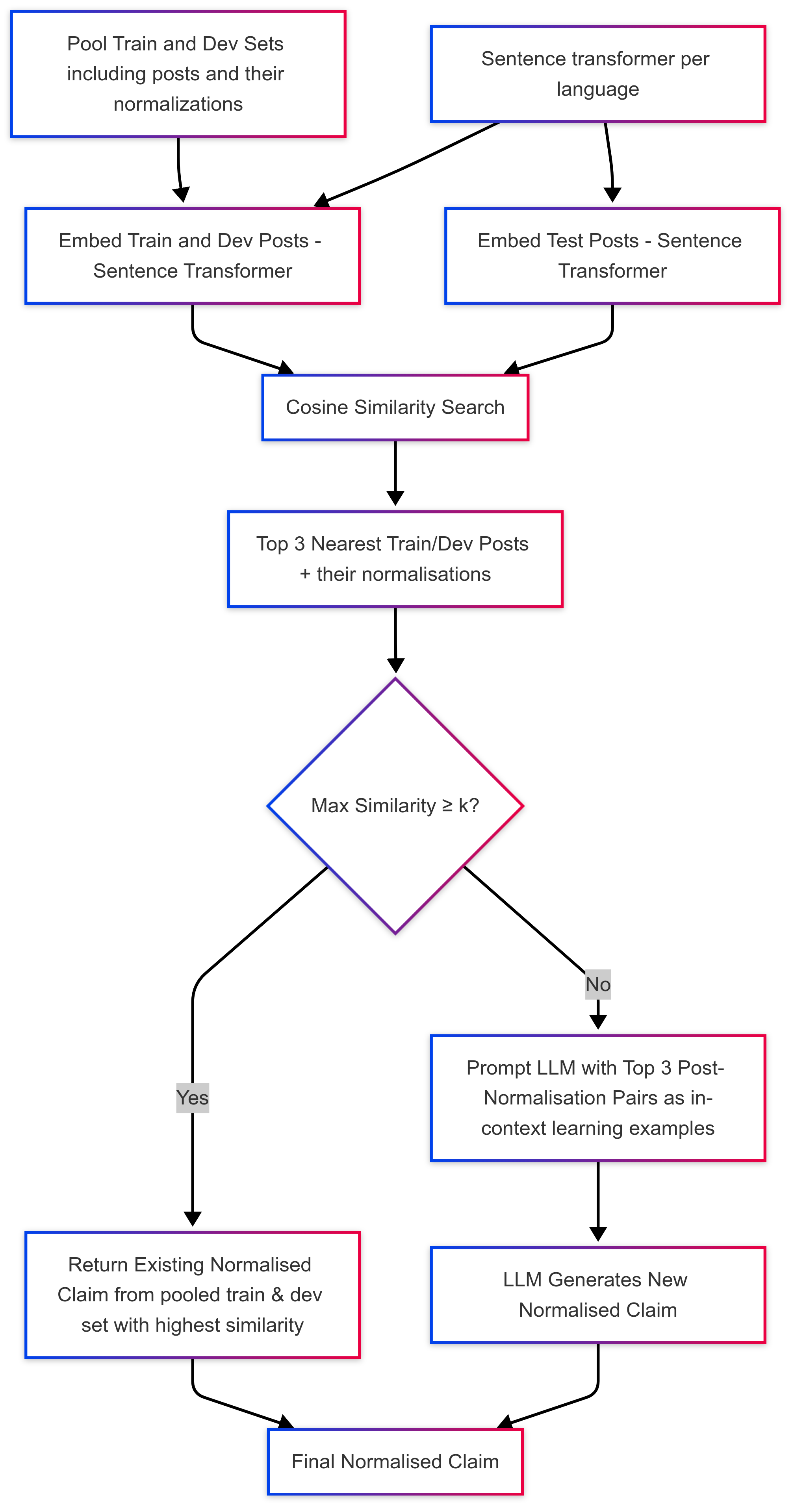} %
  \caption{Hybrid pipeline for claim normalization in monolingual settings, combining similarity-based retrieval with few-shot prompting of LLMs.}
  \label{fig:modeling}
\end{figure}

\begin{itemize} 
\item If the similarity is above a certain threshold $k$, then the existing normalized claim (from the pooled dev \& train dataset) is taken at the point of test inference. 

\item If it is below a certain threshold, then an LLM (\texttt{gpt-4o-mini}) is prompted to do the normalization, in a style similar to the CACN in \cite{sundriyal-etal-2023-chaos}. Unlike CACN, which uses static in-context examples, our approach selects the top 3 most similar instances dynamically based on cosine similarity, enabling adaptive prompting for each test post. We also instruct the LLM to produce the output in the target language (e.g. German, French, Spanish etc.)
\end{itemize}
Figure \ref{fig:modeling} illustrates the modeling procedure described above.

Note that this process of ``retrieval-first, LLM inference as a backup mechanism" for claim normalization works only in the presence of training \& dev data, i.e. for the monolingual part of \emph{CheckThat! Task 2}. As we do not have training data for the seven zero-shot language tasks, for zero-shot languages, we fall back to a simplified CACN-style prompt using a small set of static English in-context examples, regardless of the target language.  The prompt skeleton for the system message and the user message are given in Listing \ref{lst:prompt}.

\begin{lstlisting}[style=promptjson,
                   caption={Normalization prompt of system and user messages, with dynamically passed posts and lang parameters, as well as few-shot examples},
                   label={lst:prompt},
                   float,
                   floatplacement=H]
{
  "system_message": {
    "role": "system",
    "content": "You are an assistant that, given a post, identifies the central check-worthy claim contained within it. Summarize it in one sentence. Internally, you must perform detailed step-by-step reasoning to arrive at the final claim, but do not output any of your reasoning. Your final response should be a single sentence containing only the normalized claim, with no prefatory phrases such as 'the central claim is,' 'therefore,' or any similar expressions. Even if the input is ambiguous, always provide your best normalized claim without indicating that more context is needed. You will receive some examples in following ISO language code: {lang} and you will give responses in the following ISO language code: {lang}. Do not use any language other than {lang} in your response. Do not respond in English unless the post you need to normalize is in English."
  },
  "user_message": {
    "role": "user",
    "content": "Identify the central claim in the given post: {post}\nLet's think step by step."
  }
}
\end{lstlisting}
 
\section{Results and Discussion}

The results based on our approach and the test dataset are given in Table \ref{tab:hybrid_lang_results}.
\begin{table}[t]
  \centering
  \caption{Competition performance (at submission) of the hybrid retrieval–generation
           pipeline by language on the test set.
           Models marked “Monolingual’’ use language-specific sentence transformers
           (Hugging Face IDs shown).}
  \label{tab:hybrid_lang_results}

  \begin{tabularx}{\linewidth}{lcccY}
    \toprule
    \multicolumn{5}{c}{\textbf{Monolingual}}\\
    \midrule
    Lang & Threshold $(k)$ & METEOR & Rank & Sentence Transformer (HF ID)\\
    \midrule
    ara & 0.90 & 0.5035 & 2 & intfloat/multilingual-e5-base\\
    deu & 0.80 & 0.3859 & 1 & intfloat/multilingual-e5-base\\
    eng & 0.60 & 0.4521 & 2 & sentence-transformers/msmarco-distilbert-base-v3\\
    fra & 0.80 & 0.5273 & 1 & intfloat/multilingual-e5-base\\
    por & 0.90 & 0.5770 & 1 & intfloat/multilingual-e5-base\\
    spa & 0.80 & 0.6077 & 1 & intfloat/multilingual-e5-base\\
    pol & 0.80 & 0.4065 & 1 & intfloat/multilingual-e5-base\\
    hi  & 0.80 & 0.3001 & 2 & krutrim-ai-labs/Vyakyarth\\
    mr  & 0.80 & 0.2608 & 4 & krutrim-ai-labs/Vyakyarth\\
    pa  & 0.80 & 0.2567 & 8 & krutrim-ai-labs/Vyakyarth\\
    ta  & 0.80 & 0.4702 & 3 & krutrim-ai-labs/Vyakyarth\\
    tha & 0.80 & 0.5859 & 1 & intfloat/multilingual-e5-base\\
    msa & 0.80 & 0.5650 & 1 & LazarusNLP/all-indo-e5-small-v4\\
    \midrule
    \multicolumn{5}{c}{\textbf{Zero-shot}}\\
    \midrule
    ces & -- & 0.1959 & 3 & --\\
    ell & -- & 0.2250 & 4 & --\\
    kor & -- & 0.1156 & 3 & --\\
    te  & -- & 0.3171 & 5 & --\\
    bn  & -- & 0.2435 & 4 & --\\
    ron & -- & 0.2220 & 4 & --\\
    nld & -- & 0.1608 & 5 & --\\
    \bottomrule
  \end{tabularx}
\end{table}
The pipeline shows strong performance for many of the monolingual language settings. By focusing on the top of the similarity distribution and directly selecting the most similar normalized claim from the pooled dev and train sets, the system achieves strong results across several languages. Even for English, where we used a relatively low threshold of 0.6, our approach performed competitively, with only a small gap from the top-ranked submission.

The system performs best on Germanic and Romance languages, likely due to high-quality pretrained sentence transformers and greater linguistic overlap with English. Thai and Malay also show strong results, despite being typologically distinct, suggesting robustness of the multilingual embeddings. Performance is lower for the four languages from the Indo-Aryan and Dravidian families. This may reflect limitations in the underlying sentence transformers, which were not fine-tuned specifically for these language groups. Improved results could likely be achieved with better multilingual models or transformers fine-tuned on more data from these linguistic regions.

In contrast, the zero-shot setup - based solely on LLM prompting without retrieval - consistently under-performs, underscoring the need for better prompting strategies or multilingual LLM fine-tuning. 

Recall that the proposed system was motivated due to the findings in the EDA analysis, which studied the semantic similarity for many of the claims in between the test and train \& dev dasets. Naturally, the effectiveness of the retrieval system naturally depends on the degree of said similarity, which highlights both the use and the limitation of the proposed approach.

\section{Future Work}
In the presence of substantial overlap between train, dev, and test sets, training traditional seq2seq models becomes problematic due to the ease of overfitting. For future iterations of this task, a more careful stratification — ensuring minimal source overlap across splits — would be highly beneficial, making non-LLM-based learning both more realistic and more comparable.

Within the current setup we see a clear potential for improving first-stage retrieval by selecting sentence transformers better tailored to underperforming languages. This could either enable direct inference or provide more contextually relevant in-context examples to guide LLM prompting.

Additionally, the LLM prompting component itself offers room for refinement. Although we initially experimented with the detailed CACN prompt from \cite{sundriyal-etal-2023-chaos}, we found that it frequently led to outputs prefaced with auxiliary framing like “The post claims that...” — phrasing that negatively impacted the METEOR score and proved difficult to remove post hoc. This seems to stem from the inclusion of reverse check-worthiness in the prompt. While this component may offer benefits, as argued in the original CACN paper, we ultimately opted for a simpler prompt composed of only in-context learning examples. This yielded cleaner outputs without the undesired boilerplate phrasing.

\section{Conclusions}
 
This paper establishes a lightweight yet competitive framework for claim normalization in the \textit{CheckThat!~2025} Task 2.  A thorough EDA on the English split revealed two systemic obstacles: (i)
low semantic overlap between raw posts and the normalizations created by humans (ii) substantial claim overlap (to the extent of even duplication) across train, dev, and test sets.  These insights motivated a \emph{retrieval-first, LLM-backed} pipeline. For every test post we retrieve its nearest neighbour from the pooled train–dev corpus. If the similarity exceeds a language-specific threshold, the neighbour’s known normalization is reused, otherwise a simple adaptive in-context prompt (taking the top 3 nearest neighbour's post-normalizations pairs) is fed to \texttt{gpt-4o-mini}.  The approach avoids training a seq2seq model that would likely overfit the leaked validation data, and at the same time struggle on low-context posts.

For the test submissions, the system ranks near the top for many monolingual tracks, especially Germanic and Romance languages, while still delivering solid results for Thai and Malay.  Lower scores for the Indo-Aryan and the Dravidian languages suggest room for a stronger, language-tailored, sentence transformers and more diverse in-context exemplars.  

In the zero-shot setting, pure LLM normalization performs worse, which shows the need to improve the prompting procedure in cases where training data is not available.

\begin{acknowledgments}
We thank the DS@GT CLEF team for providing valuable comments and suggestions.
This research was supported in part through research cyberinfrastructure resources and services provided by the Partnership for an Advanced Computing Environment (PACE) at the Georgia Institute of Technology, Atlanta, Georgia, USA.
\end{acknowledgments}

\section*{Declaration on Generative AI}
During the preparation of this work, the authors used OpenAI GPT-4o: Grammar and spelling check. After using this tool, the authors reviewed and edited the content as needed and take full responsibility for the publication’s content.

\bibliography{main}

\end{document}